\documentclass{article}

\PassOptionsToPackage{numbers, compress}{natbib}
\usepackage[preprint]{neurips_2020}

\usepackage[utf8]{inputenc} % allow utf-8 input
\usepackage[T1]{fontenc}    % use 8-bit T1 fonts
\usepackage{hyperref}       % hyperlinks
\usepackage{url}            % simple URL typesetting
\usepackage{booktabs}       % professional-quality tables
\usepackage{amsfonts}       % blackboard math symbols
\usepackage{nicefrac}       % compact symbols for 1/2, etc.
\usepackage{amsmath}
\usepackage{amsthm}
\usepackage{graphicx}
\usepackage{algorithm}
\usepackage{algorithmic}
\usepackage{color}
\usepackage{colortbl}
\usepackage{comment}
\usepackage{wrapfig}

\newcolumntype{P}[1]{>{\centering\arraybackslash}p{#1}}

\usepackage{amssymb}
\usepackage{enumitem}
\usepackage{array}
\usepackage{multirow}
\usepackage{bm}
\usepackage{url}
\usepackage{subcaption}

\theoremstyle{remark}
\newtheorem{remark}{Remark}
\newtheorem{theorem}{Theorem}

\title{Eigen-GNN: A Graph Structure Preserving Plug-in for GNNs}
\author{%
  Ziwei Zhang \\
  Tsinghua University \\
  \texttt{zw-zhang16@mails.tsinghua.edu.cn} \\
     \And
   Peng Cui \\
   Tsinghua University \\
   \texttt{cuip@tsinghua.edu.cn} \\
   \And
   Jian Pei \\
   Simon Fraser University \\
   \texttt{jpei@cs.sfu.ca} \\
   \And
   Xin Wang \\
   Tsinghua University \\
   \texttt{xin\_wang@tsinghua.edu.cn} \\
   \And
   Wenwu Zhu \\
   Tsinghua University \\
   \texttt{wwzhu@tsinghua.edu.cn} \\
}

\begin{document}
\begin{sloppy}

\maketitle
\vspace{-0.5cm}
\begin{abstract}
    Graph Neural Networks (GNNs) are emerging machine learning models on graphs. Although sufficiently deep GNNs are shown theoretically capable of fully preserving graph structures, most existing GNN models in practice are shallow and essentially feature-centric. We show empirically and analytically that the existing shallow GNNs cannot preserve graph structures well. To overcome this fundamental challenge, we propose Eigen-GNN, a simple yet effective and general plug-in module to boost GNNs ability in preserving graph structures.
    Specifically, we integrate the eigenspace of graph structures with GNNs by treating GNNs as a type of dimensionality reduction and expanding the initial dimensionality reduction bases. Without needing to increase depths, Eigen-GNN possesses more flexibilities in handling both feature-driven and structure-driven tasks since the initial bases contain both node features and graph structures. We present extensive experimental results to demonstrate the effectiveness of Eigen-GNN for tasks including node classification, link prediction, and graph isomorphism tests.
\end{abstract}
\vspace{-0.5cm}
\section{Introduction}
Graphs are natural representations for complex data such as social networks, biomedical graphs, and traffic networks.
Besides carrying relation information through graph structures, graphs are often associated with rich content information such as attributes of nodes. Content (features) and structures often provide information complementary to each other.  Some analytics tasks focus on the content information, e.g., in document topic classifications, the content of documents usually provides dominant information. We call such tasks \emph{feature-driven}. In some other analytics tasks, structures are the major player.  A great example of such \emph{structure-driven tasks} is influence analysis in social networks. Of course, there are always some analytics tasks where both content and structure information are needed. For example, in social recommendations, both user profiles (content) and user interactions (structure) are indispensable in understanding user preferences.

Recently, Graph Neural Networks (GNNs)%\footnote{Some studies also adopt the name Graph Convolutional Networks. We use the two terms interchangeably.}
are emerging machine learning models on graphs, and are expected to provide a unified framework to deal with features and structures simultaneously. For example, in the message-passing framework~\cite{gilmer2017neural}, nodes exchange information with their neighbors in each message-passing step to update their feature information. In this way, GNNs model node attributes and graph structures in an end-to-end learning architecture.

It has been proven theoretically that GNNs with a sufficiently large number of layers can fully preserve many important graph structures such as the limiting distribution of a random walk on graphs~\cite{xu2018representation,klicpera2019predict} and graph moments of any order~\cite{dehmamy2019understanding}. However, training deep GNNs suffers from many practical challenges, such as over-smoothing~\cite{li2018deeper,wu2019comprehensive}. In practice most successful GNNs are shallow, having no more than three or four layers~\cite{kipf2017semi,zhang2018deep}.

However, shallow GNNs are distant from ideal GNNs that have a large number of layers.  Recent analysis shows that the existing shallow GNNs essentially are feature-centric, i.e., node attributes play major roles, and graph structures only provide auxiliary information. For example, Li~\textit{\textit{et~al.}}~\cite{li2018deeper} analyzed GNNs as a special form of Laplacian smoothing of node attributes. Maehara~\cite{maehara2019revisiting} and Wu~\textit{\textit{et~al.}}~\cite{wu2019simplifying} showed that GNNs are equivalent to a low-pass filter by treating node features as graph signals. Given these discussions showing the strength of GCNs in preserving features, a critical next question is \emph{whether the shallow GNNs in practice can sufficiently preserve graph structures}, which motivates this study.

To answer this question, we first report experimental analysis on a series of synthetic datasets (please refer to Section~\ref{sec:preli} for details). We observe consistent results with the aforementioned analysis: in the structure-driven tasks, the existing shallow GNNs have poor performance. We further examine this observation by treating GNNs as a type of dimensionality reduction process. We show that the features of nodes provide the initial bases for the dimensionality reduction, making the resulted predictions of the existing shallow GNNs tend to be feature-centric. Therefore, the existing shallow GNNs are incapable of sufficiently preserving graph structures in practice.

\emph{Can we have a simple and general mechanism to empower the practical shallow GNNs to preserve graph structures well?} To tackle this fundamental challenge, we propose Eigen-GNN\footnote{We will release the source code once the paper is accepted for publication.}, a simple yet effective and general plug-in module to boost GNNs ability in preserving graph structures. Specifically, we integrate the eigenspace of graph structures with GNNs by concatenating the eigenvectors of a graph structure matrix to the node attributes. In this way,
since the initial bases are expanded to contain both node features and graph structures, Eigen-GNN has dramatically enhanced capabilities in exploring node features and graph structures simultaneously.
We also demonstrate that Eigen-GNN has several desirable theoretical properties such as permutation-equivariance and generality in plugging into many existing GNNs.

We conduct extensive experiments for tasks including node classification, link prediction, and graph isomorphism tests. The experimental results clearly show that our proposed method consistently and significantly outperforms the baselines when the tasks and datasets are more structure-driven, and retains comparable performance with existing GNNs in feature-driven scenarios.

Our contributions in this paper are summarized as follows:
\begin{itemize}[leftmargin = 0.6cm]
\item We demonstrate that most existing GNNs are shallow and cannot preserve graph structures well in practice through both empirical analysis and analytical exploration.
\item We propose Eigen-GNN, a simple yet effective and general plug-in module to boost GNNs ability in preserving graph structures. Eigen-GNN has several desirable theoretical properties and can be applied to many existing GNN architectures.
\item Our extensive experimental results demonstrate that the proposed Eigen-GNN can preserve both features and graph structures more effectively and flexibly.
\end{itemize}

\section{Related Work}\label{sec:ref}
We briefly review related works in GNNs and refer readers to~\cite{wu2019comprehensive,zhang2018deep,zhou2018graph} for comprehensive surveys.

Bruna~\textit{et~al.}~\cite{bruna2014spectral} first defined graph convolutions using graph signal processing~\cite{shuman2013emerging}.
Kipf and Welling~\cite{kipf2017semi} simplified the graph convolution using the first-order polynomial, i.e., only considering the immediate neighbors of nodes. MPNNs~\cite{gilmer2017neural} and GraphSAGE~\cite{hamilton2017inductive} unify these methods using a ``message-passing'' framework, i.e., nodes aggregate information from their neighborhoods as messages. Later studies usually follow these frameworks by proposing different variants.

To understand the effectiveness of GNNs, Li~\textit{et~al.}~\cite{li2018deeper} showed that GNNs are a special form of Laplacian smoothing.
Hou~\textit{et~al.}~\cite{hou2020measuring} further proposed a metric to measure the smoothness of node features and node labels.
Wu~\textit{et~al.}~\cite{wu2019simplifying} showed that the existing GNNs are equivalent to a fixed low-pass filter of graph signals and proposed an extremely simplified GNN by removing all the non-linearities. Maehara~\cite{maehara2019revisiting} took a similar idea and showed that adding an extra Multi-Layer Perceptron (MLP) layer can further increase the non-linear manifold learning capability of GNNs. Kipf and Welling~\cite{kipf2017semi}, Zhang~\textit{et~al.}~\cite{zhang2018end}, and Xu~\textit{et al.}~\cite{xu2019powerful} considered the connection between GNNs and the Weisfeiler-Lehman (WL) kernel for graph isomorphism tests. Dehmamy~\textit{et~al.}~\cite{dehmamy2019understanding} showed that GNNs with an infinite number of layers can preserve graph moments of any order, the statistics that characterize the random process of graph generation. However, whether shallow GNNs can preserve graph structures well in practice remains an open problem.

There are also recent attempts in increasing the depth of GNNs~\cite{li2019deepgcns,rong2019dropedge,zhao2019pairnorm}, which is orthogonal to the study of this paper.

\section{How Well Can Shallow GNNs Preserve Graph Structures?}\label{sec:observations}

\subsection{An Empirical Study}\label{sec:preli}
To manifest the capability of shallow GNNs in structure-driven and feature-driven tasks, we first conduct some experiments on synthetic datasets.

\paragraph{Datasets Generation and Methods in Comparison}\label{sec:preli1}
We generate synthetic datasets that contain two components: graph structures and node features. For graph structures, we partition nodes into $l$ $(l >0)$ balanced communities and construct edges using the Stochastic Blockmodel~\cite{airoldi2008mixed}. The nodes within the same community have a high probability of forming edges and those in different communities have a low probability of forming edges. We use the id of the community (a positive integer between $1$ and $l$) that a node belongs to as the structure-driven label $c_{\text{struc}}$ of the node.

For node features, we randomly divide nodes into $l$ balanced groups.  We generate a random vector for each group, called the \emph{group vector}. The features of a node are generated following a Gaussian distribution with the mean being the group vector of the group that the node belongs to.  The group id (also a positive integer between $1$ and $l$) is used as the feature-driven label $c_{\text{feat}}$ of the node.

The final node label follows a Bernoulli distribution: $c = c_{\text{struc}}$ with probability $\gamma$ and $c = c_{\text{feat}}$ with probability $1 - \gamma$, where $0 \leq \gamma \leq 1$ is a parameter controlling the degree to which the node label prediction task is structure- or feature-driven. We call all the nodes carrying the same label as a \emph{class}. Among all the nodes in class $i$ $(1 \leq i \leq l)$, some are assigned the label due to the structure and the others are assigned the label due to and are manifested by the features. As two extremes, when $\gamma = 1$, the node label prediction task is completely structure-driven and, when $\gamma = 0$, the task is completely feature-driven.
More details about the synthetic datasets can be found in Appendix~\ref{sec:additional-details-syn}. We compare three different methods:
\begin{itemize}[leftmargin = 0.6cm]
\item $\text{GCN}X_{\text{feature}}$: this is the original GCN in~\cite{kipf2017semi} taking features as inputs, where parameter $X$ indicates the number of hidden layers in the GCN. We test GCNs with $1$, $2$, $3$, and $5$ layers.
\item $\text{MLP}_{\text{feature}}$: we use a two-layer Multi-Layer Perceptron on node features~\cite{kipf2017semi}, i.e., a neural network with two fully connected layers. $\text{MLP}_{\text{feature}}$ does not learn any graph structure.
\item DeepWalk~\cite{perozzi2014deepwalk}: a network embedding method to learn node representations and preserve graph structures. No node feature is used. We add a fully connected layer and a softmax layer on the learned embedding vectors for classification.
\end{itemize}
More experimental settings are provided in Appendix~\ref{sec:additional-details-syn}.
In this section, please ignore the curves of $\text{Eigen-GCN}$ in Figures~\ref{fig:pre} and~\ref{fig:pre2}, which will be discussed later in Section~\ref{sec:revisit}.

\paragraph{Observations.} First, we consider the extreme case where only graph structures are useful in the label prediction, i.e., $\gamma = 1$. In this case, to perform well, a model has to learn sufficient information about graph structures. Figure~\ref{fig:pre} shows the results. We have the following findings.
\begin{itemize}[leftmargin = 0.5cm]
\item The accuracy of $\text{MLP}_{\text{feature}}$ is about $10\%$. Since there are 10 balanced classes, this accuracy is roughly the same as random guessing. This verifies that node features indeed are not useful here.
\item GCNs outperform $\text{MLP}_{\text{feature}}$, indicating that the existing GCNs can extract and exploit some information from the graph structures. These findings are consistent with the literature~\cite{kipf2017semi}.
\item Increasing the number of hidden layers in GCNs from $1$ to $3$ improves the accuracy. This verifies that deeper GCNs have a better capability in preserving graph structures. However, when GCNs have more layers, e.g., $\text{GCN}5_{\text{feature}}$, the performance tends to saturate or even drop (though residual connections are added), showing that
    training deep GCNs has unsolved practical challenges.
\item DeepWalk outperforms all the existing GCNs. This illustrates the weakness of GCNs in preserving graph structures. DeepWalk conducts random walks and takes the skip-gram model~\cite{mikolov2013distributed} to explicitly preserve graph structures. GCNs only utilize graph structures in aggregating node neighborhoods. The insufficiency of learning and preserving graph structures in GCNs explains the inferior performance of GCNs in structure-driven tasks.
\end{itemize}

Next, we vary $\gamma$ to mimic different kinds of tasks. Recall that the larger $\gamma$, the more structure-driven a task, and vice versa. The results are shown in Figure~\ref{fig:pre2}. We have the following observations.
\begin{itemize}[leftmargin = 0.6cm]
\item When $\gamma$ approaches 1, the results are consistent with those in Figure~\ref{fig:pre}. DeepWalk preserves graph structures better than GCNs. $\text{MLP}_{\text{feature}}$ gets the worst results since it does not use any graph structural information.
\item When $\gamma$ approaches 0, i.e., the task is heavily feature-driven, $\text{MLP}_{\text{feature}}$ achieves the best results. GCNs achieve inferior performance as they are misled to some extent by graph structures. DeepWalk performs poorly, the performance being similar to random guess, since it does not utilize any feature information.
\item No existing method can perform well with respect to various $\gamma$ values.  This clearly indicates that the existing models cannot preserve features and structure information well simultaneously.

\end{itemize}
\begin{figure}[t]
\begin{subfigure}{.5\textwidth}
  \centering
  \includegraphics[height=3.2cm]{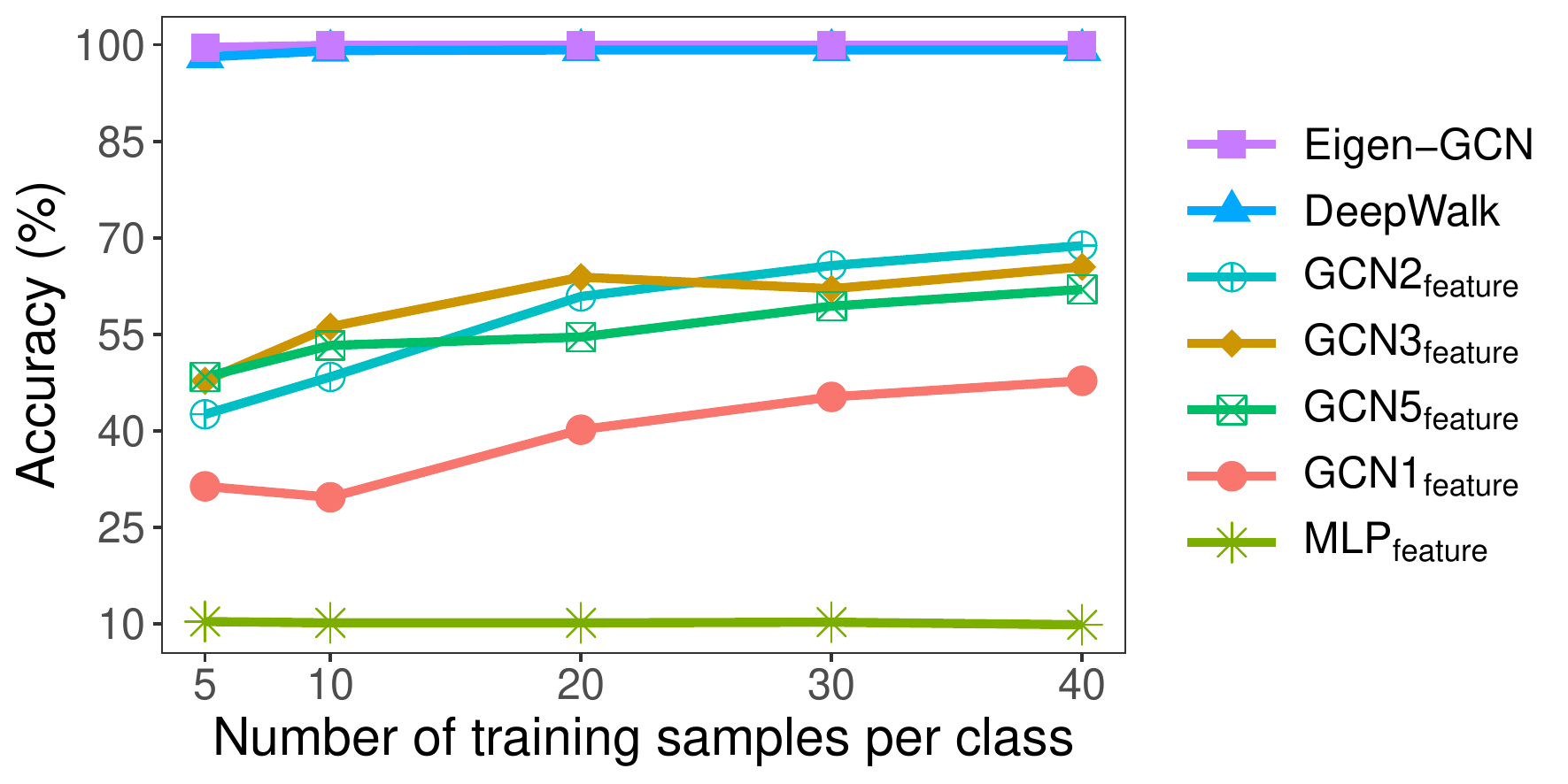}\\
  \caption{}
  \label{fig:pre}
\end{subfigure}
\begin{subfigure}{.5\textwidth}
  \centering
  \includegraphics[height=3.2cm]{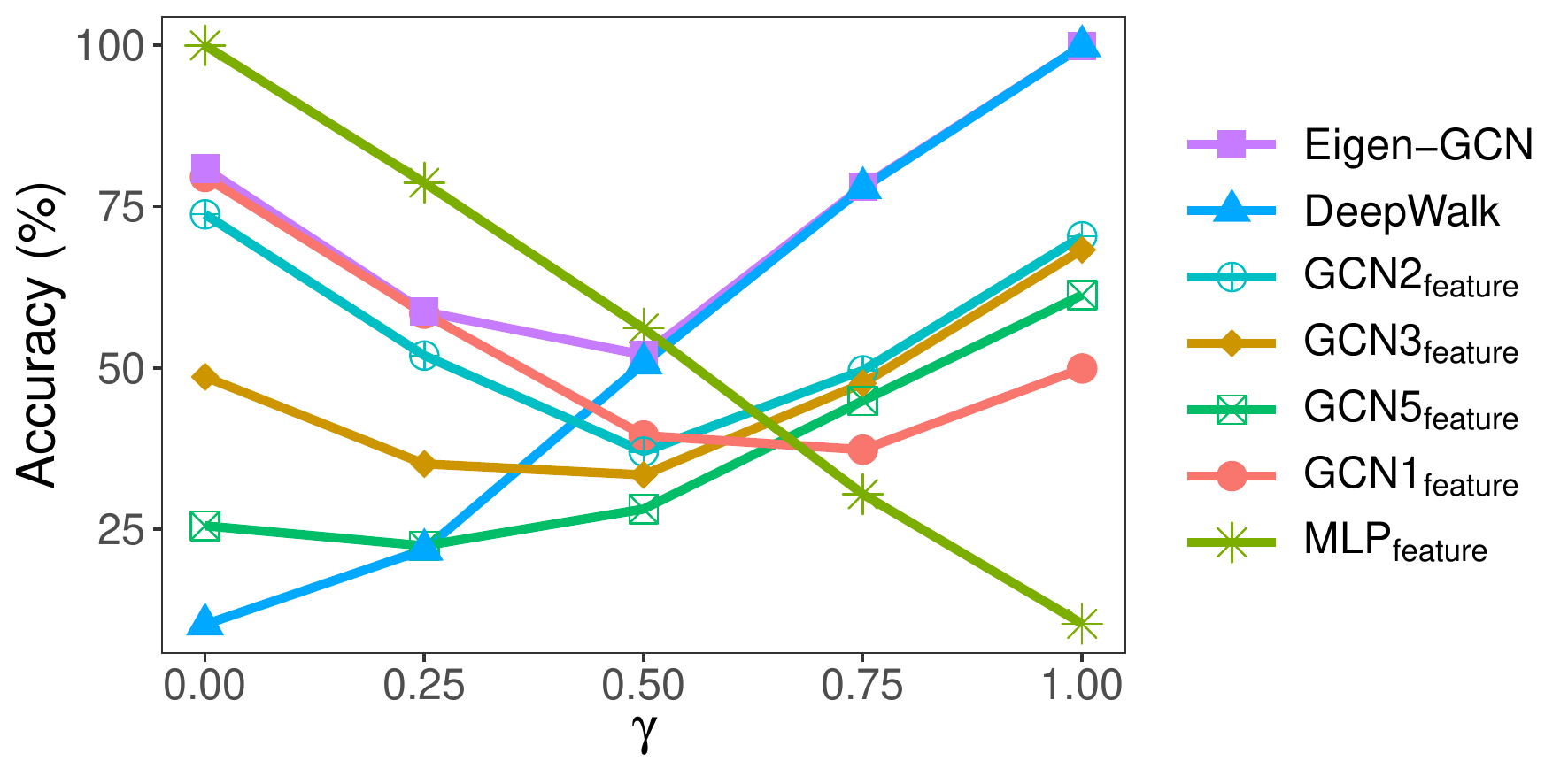}\\
  \caption{}
  \label{fig:pre2}
\end{subfigure}
\caption{The experimental results on synthetic datasets (a) in a completely structure-driven task, i.e., $\gamma = 1$, (b) when varying $\gamma$ between $0$ and $1$. Best viewed in color. }
\vspace{-0.15cm}
\end{figure}
In summary, the experimental results on synthetic datasets clearly show that the existing shallow GNNs cannot preserve graph structures well in practice. Indeed, no existing method can be consistently competent in both structure- and feature-driven tasks.
Next, to better understand this phenomenon, we provide an analytical investigation using dimensionality reduction.

\subsection{GNNs as Dimensionality Reduction}\label{sec:dimreduction}

Consider a graph $\pmb{G}=(\mathcal{V},\mathcal{E},\mathbf{X})$, where $\mathcal{V}=\left\{v_1,...,v_N\right\}$ is a set of $N$ nodes, $\mathcal{E} \subseteq \mathcal{V} \times \mathcal{V}$ is a set of $M$ edges, and $\mathbf{X} \in \mathbb{R}^{N \times f}$ is an optional node feature matrix where $f$ is the number of features. Denote by $\mathbf{A} \in \mathbb{R}^{N \times N}$ the adjacency matrix, and by $\mathbf{A}_{i,:}$, $\mathbf{A}_{:,j}$, and $\mathbf{A}_{i,j}$, respectively, the $i^{th}$ row, the $j^{th}$ column and an element in the matrix. We assume connected and undirected graphs, i.e., $\mathbf{A}_{i,j} = \mathbf{A}_{j,i}, 1 \leq i,j \leq N$. We use bold uppercases (e.g., $\mathbf{Z}$) and bold lowercases (e.g., $\mathbf{z}$) to denote matrices and vectors, respectively. Functions are marked by curlicue, e.g., $\mathcal{F}(\cdot)$.  We denote a non-linear activation function such as sigmoid or ReLU as $\sigma(\cdot)$.

Our analysis starts with the observation that many existing GNNs can be unified into the following framework. Denote by $\mathbf{H}^{(l)} \in \mathbb{R}^{N\times d_l}$ the representations of the nodes in the $l^{th}$ hidden layer, where $d_l$ is the dimensionality of layer $l$ and $\mathbf{H}^{(0)} = \mathbf{X}$ are the input features, by $\mathbf{W}^{(l)} \in \mathbb{R}^{d_l \times d_{l+1}}$ the parameters, and by $\mathcal{F}(\mathbf{A}) \in \mathbb{R}^{N \times N}$ a function on the graph structure. The $(l+1)^{th}$ layer in a GNN can be formulated as:
\begin{small}
\begin{equation}\label{eq:GCN1}
 \mathbf{H}^{(l+1)} = \sigma\left( \mathcal{F}\left(\mathbf{A}\right)\mathbf{H}^{(l)} \mathbf{W}^{(l)} \right).
\end{equation}
\end{small}
For example, a well-known GNN variant by Kipf and Welling~\cite{kipf2017semi} adopts the following function:
\begin{small}
\begin{equation}\label{eq:GCN2}
\mathcal{F}\left( \mathbf{A} \right) = \tilde{\mathbf{D}}^{-\frac{1}{2}} \tilde{\mathbf{A}} \tilde{\mathbf{D}}^{-\frac{1}{2}},
\end{equation}
\end{small}
where $\tilde{\mathbf{A}} = \mathbf{A} + \mathbf{I}_N$, $\mathbf{I}_N$ is the identity matrix, and $\tilde{\mathbf{D}}_{i,i} = \sum_j \tilde{\mathbf{A}}_{i,j}$ is the diagonal degree matrix. We list several other GNNs and their corresponding $\mathcal{F}\left(\mathbf{A} \right)$ in Table~\ref{tab:framework} in the Appendix.

Denote by $\mathbf{F} = \mathcal{F}\left(\mathbf{A}\right)$. $\mathbf{F}$ actually encodes the raw structure information of the graph. For example, using Eq.~\eqref{eq:GCN2}, we have
\begin{small}
\begin{equation}
    \mathbf{F}_{i,j} = (\tilde{\mathbf{D}}_{i,i}\tilde{\mathbf{D}}_{j,j})^{-\frac{1}{2}} \tilde{\mathbf{A}}_{i,j}.
\end{equation}
\end{small}That is, $\mathbf{F}_{i,:}$ is a normalized adjacent vector of node $v_i$, encoding the second-order proximity between nodes~\cite{wang2016structural}. In DCNN~\cite{atwood2016diffusion} and PPNP~\cite{klicpera2019predict}, $\mathbf{F}$ encodes the transition probability between nodes. Eq.~\eqref{eq:GCN1} can be interpreted as a three-step dimensionality reduction process by executing the calculation from left to right:
\begin{itemize}[leftmargin = 0.6cm]
\item Step 1: $\mathbf{F}^\prime = \mathbf{F}\mathbf{H}^{(l)} $, i.e., projecting $\mathbf{F}$ into a subspace spanned by $\mathbf{H}^{(l)}$ to obtain $\mathbf{F}^\prime$.
\item Step 2: $\mathbf{F}^\prime$ is further transformed by a linear mapping $\mathbf{W}^{(l)}$ followed by a non-linear function $\sigma(\cdot)$, i.e., $\mathbf{H}^{(l+1)}=\sigma\left(\mathbf{F}^\prime \mathbf{W}^{(l)}\right)$ as refined low-dimensional representations.
\item Step 3: repeat the above two steps using $\mathbf{H}^{(l+1)}$ as the new base in Step 1.
\end{itemize}

\begin{remark}
GNNs can be regarded as a (non-linear) dimensionality reduction procedure with each GNN layer performing one dimensionality reduction process. The node features provide the initial bases for the dimensionality reduction.
\end{remark}

Since most of the existing GNNs in practice are shallow and the number of iterations in the dimensionality reduction is determined by the number of layers, the initial bases for dimensionality reduction play crucial roles and provide important inductive biases for the GNNs. If the initial bases are solely determined by node features as in the existing shallow GNNs, the resulted models are inevitably feature-centric and cannot well preserve graph structures.

In addition, the existing GNNs are struggling to handle the situations when no node feature is available. A commonly used trick is to use a one-hot encoding of node IDs~\cite{kipf2017semi,murphy19a}, i.e., setting $\mathbf{X} = \mathbf{I}_N$. However, using a one-hot encoding will dramatically increase the number of parameters and make the model unable to retain permutation-equivariance~\cite{murphy19a}.
Another heuristic method is to use node degrees as node features~\cite{xu2019powerful}, but it can only encode limited graph structure information.

\section{Eigen-GNN}\label{sec:method}

\paragraph{The Model}

As analyzed in Section~\ref{sec:dimreduction}, the main reason that the existing shallow GNNs fail to preserve graph structures well is that the initial dimensionality reduction bases, $\mathbf{H}^{(0)} = \mathbf{X}$, are completely biased to features only and do not contain any structure information. To fix the problem, we need to find a suitable subspace where useful graph structure information can be preserved. It is well known in spectral graph theory~\cite{chung1997spectral} that the eigenspace of a graph provides informative low-dimensional spaces regarding graph structures. For example, spectral clustering~\cite{ng2002spectral} adopts the top-$d$ smallest eigenvectors of the Laplacian matrix for node clustering, and network embedding adopts the top-$d$ largest eigenvectors of a polynomial function of the adjacency matrix for unsupervised node representation learning~\cite{zhang2018arbitrary}. Inspired by those successes, our idea is to integrate the eigenspace of graph structures with GNNs by expanding the initial dimensionality reduction bases.

To keep it simple and general, we expand the initial dimensionality reduction bases by directly concatenating the top-$d$ eigenvectors of a graph structure matrix with node features, i.e., we set
\begin{small}
\begin{equation}\label{eq:eigenGCN}
    \mathbf{H}^{(0)} = \left[ \mathbf{X}, f\left(\mathbf{Q}\right)\right],
\end{equation}
\end{small}
where $\mathbf{X}$ is the feature matrix, $\mathbf{Q} \in \mathbb{R}^{N \times d}$ are the top-$d$ eigenvectors corresponding to the largest absolute eigenvalues of a specified matrix $\mathcal{G}(\mathbf{A})$ of the graph structure, $f(\cdot)$ is a simple function such as identity mapping or a normalization operator, and $\left[\cdot ,\cdot \right]$ is the concatenation operator. In this paper, we reuse the symmetrically normalized adjacency matrix in Eq.~\eqref{eq:GCN2} as the graph structure matrix, i.e., set $\mathcal{G}(\mathbf{A}) =\tilde{\mathbf{D}}^{-\frac{1}{2}} \tilde{\mathbf{A}} \tilde{\mathbf{D}}^{-\frac{1}{2}}$, but our method can be easily generalized to other matrices, such as the Laplacian matrix or the transition matrix.

Rather than being a new GNN architecture, our proposed Eigen-GNN can be used as a plug-in module to enhance the capability of many existing GNNs in preserving graph structures. As both node features and graph structure information are captured in the initial dimensionality reduction bases, Eigen-GNN is flexible and adaptive in handling both structure-driven and feature-driven tasks. Moreover, as the eigenspace is independent of node attributes, Eigen-GNN can easily handle featureless graphs by only using the eigenbasis, which is in contrast with the existing GNNs that can only use heuristics such as a one-hot encoding or degrees.

As Eigen-GNN only provides the initial dimensionality reduction bases, it can work jointly with different GNNs, including those designed for signed or multi-relational graphs, like propagating between positive/negative edges~\cite{derr2018signed} and learning different weights for different edge types~\cite{schlichtkrull2018modeling}.

\paragraph{Several Desirable Properties of Eigen-GNN}

Firstly, we prove that a specific Eigen-GNN variant is permutation-equivariant, i.e., the model is equivariant under graph isomorphisms (permutation of node IDs), as long as the top-$d$ eigenvalues of $\mathcal{G}(\mathbf{A})$ are unique.

\begin{theorem}\label{thm1}
For two graphs $\pmb{G}=(\mathcal{V},\mathcal{E},\mathbf{X})$ and $\pmb{G}^\prime=(\mathcal{V}^\prime,\mathcal{E}^\prime,\mathbf{X}^\prime)$, denote by $\mathbf{H}^{(l)},\mathbf{U}^{(l)}, 0 \leq l \leq L$, respectively, the representations of $\mathcal{V}$ and $\mathcal{V}^\prime$ in the $l^{th}$ hidden layer of an Eigen-GNN. We assume the top-$d$ eigenvalues of $\mathcal{G}(\mathbf{A})$ are unique for $\pmb{G}$ and $\pmb{G}^\prime$ and use $f(x) = \left|x \right|$. Then, if there exists a bijective mapping $\mathcal{B}: \mathcal{V} \rightarrow \mathcal{V}^\prime$ so that $\mathcal{E}(i,j) = \mathcal{E}^\prime\left(\mathcal{B}(i),\mathcal{B}(j)\right),\mathbf{X}_{i,:} = \mathbf{X}^\prime_{\mathcal{B}(i),:}, \forall 1 \leq i,j \leq N$, then, $\mathbf{H}^{(l)}_{i,:} = \mathbf{U} ^{(l)}_{\mathcal{B}(i),:}, \forall 1 \leq i \leq N, \forall 0 \leq l \leq L$.
\end{theorem}
The proof is given in Appendix~\ref{sec:proofthm1}.
Permutation-equivariance is a necessary condition for intermediate layers of a permutation-invariant GNN~\cite{maron2019invariant} targeting graph-level tasks such as graph classification. By satisfying permutation-equivariance, Eigen-GNN can be applied to graph-level tasks. We further analyze this property empirically in Section \ref{sec:CSLex}.

In addition, Eigen-GNN is scalable to large graphs since we only calculate the eigenvectors corresponding to the largest absolute eigenvalues. Specifically, we have the following result.
\begin{theorem}\label{thm:complex}
The time complexity of calculating the eigenbasis of Eigen-GNN in Eq.~\eqref{eq:eigenGCN} is $O\left( T\left(M_\mathcal{G}d + Nd^2\right) \right)$, where $M_\mathcal{G}$ is the number of non-zero elements in $\mathcal{G}(\mathbf{A})$, $N$ is the number of nodes, $d$ is the preset dimensionality, and $T$ is the number of iterations (a constant).
\end{theorem}
\begin{proof}
The theorem is due to the iterative algorithms such as~\cite{lehoucq1996deflation}.
\end{proof}
The theorem shows that the time complexity mainly depends on the number of non-zero elements in the graph structure matrix $\mathcal{G}(\mathbf{A})$. By setting $\mathcal{G}(\mathbf{A})$ as a sparse matrix, e.g., the normalized adjacency matrix $\mathcal{G}(\mathbf{A}) =\tilde{\mathbf{D}}^{-\frac{1}{2}} \tilde{\mathbf{A}} \tilde{\mathbf{D}}^{-\frac{1}{2}}$, we have $M_\mathcal{G} \approx M$. In such a case, the time complexity of calculating the eigenbasis is linear with respect to the number of nodes and that of edges in the graph. Since this time complexity is on the same scale as the existing GNNs, Eigen-GNN does not incur any extra cost in scalability. We empirically verify this result in Appendix~\ref{sec:time}.

Finally, we show an interesting connection between Eigen-GNN and Simple Graph Convolution (SGC)~\cite{wu2019simplifying}, a simplified GNN variant without non-linearities.
\begin{theorem}\label{thm2}
For a graph that is not bipartite, an SGC with an infinite number of layers converges to Eigen-GNN with no hidden layer and the eigenspace dimensionality $d=1$.
\end{theorem}
The proof is given in Appendix~\ref{sec:proofthm2}. The theorem implies that, instead of integrating graph structures gradually in each layer as SGC, Eigen-GNN can directly provide the final graph structure information used by SGC using a ``short-cut'' by the eigenspace.

\section{Experimental Results}
Since Eigen-GNN is a general plug-in to enhance existing GNNs rather than a new architecture, we conduct a series of experiments to answer the following three questions. \textbf{Q1}: Can Eigen-GNN improve GNNs in structure-driven tasks? Does Eigen-GNN impair feature-driven tasks? \textbf{Q2}: Can Eigen-GNN be easily plugged into various GNN models? \textbf{Q3}: Can we empirically verify the desirable properties of Eigen-GNN in applications?

\subsection{Baselines and Experimental Settings}\label{sec:setting}
We compare the following three methods:
\begin{itemize}[leftmargin = 0.6cm]
\item $\text{GNN}_{\text{feat}}$: we report the original results of the GNN model with node features as inputs.
\item $\text{GNN}_{\text{feat+DW}}$: we run DeepWalk~\cite{perozzi2014deepwalk} on graph structures and concatenate the generated embedding vectors with node features as inputs to GNNs. This is a heuristic approach to enhance the capability of GNNs in preserving graph structures~\cite{qiu2018deepinf}.
\item  $\text{Eigen-GNN}_{\text{feat+struc}}$: our proposed method, i.e., we concatenate the eigenbasis with node features as inputs to GNNs.
\end{itemize}
We further include five methods without using node features.
\begin{itemize}[leftmargin=0.6cm]
\item $\text{GNN}_{\text{one-hot}}$: we use a one-hot encoding of node ID as inputs to GNNs~\cite{murphy19a}.
\item $\text{GNN}_{\text{degree}}$: we use a one-hot encoding of node degrees as inputs to GNNs, which is proven useful in chemistry graphs~\cite{xu2019powerful}.
\item $\text{GNN}_{\text{random}}$: we generate random features following a Gaussian distribution as inputs to GNNs.
\item $\text{GNN}_{\text{DW}}$: we use the embedding vectors of DeepWalk as inputs to GNNs.
\item $\text{Eigen-GNN}_{\text{struc}}$: we adopt the eigenbasis as the inputs to GNNs.
\end{itemize}

Although Eigen-GNN can be generally plugged into many different GNNs, it is infeasible to compare every possible GNN architecture due to the vast literature. Instead, we adopt the most prominent GNNs for the tasks as showcases (the exact model will be given in each subsection). We further clarify the adopted architecture by replacing the ``GNN'' in method names with the exact model, e.g., Eigen-GCN if we use GCN and Eigen-GAT if we use GAT.
For hyper-parameters, we search the dimensionality $d$ of the eigenspace from $\left\{32,64,128,256\right\}$. We repeat all experiments 10 times.
Additional hyper-parameters and details for reproducibility can be found in Appendix~\ref{sec:additional-details}.

\subsection{Node Classification}\label{sec:nc}

\paragraph{Revisiting the Empirical Study in Section~\ref{sec:preli}}\label{sec:revisit}
Now let us examine the performance of Eigen-GCN (we use GCN the base GNN architecture) in Figures~\ref{fig:pre} and~\ref{fig:pre2}.

\begin{itemize}[leftmargin = 0.6cm]
\item When the task is structure-driven (i.e., $\gamma$ approaches 1), Eigen-GCN achieves the best performance. This shows that our plugged-in module can empower GCNs to better capture graph structure information.
\item Eigen-GCN achieves the most stable performance with respect to $\gamma$ varying between $0$ and $1$.  This demonstrates that Eigen-GCN can handle both feature-driven and structure-driven tasks. Eigen-GCN consistently outperforms the existing GCNs when $\gamma \geq 0.5$ and retains comparable results when $\gamma < 0.5$, showing that Eigen-GCN is robust and thus a reliable choice even when the type of the tasks is unknown.
\item When the task is feature-driven (i.e., $\gamma$ approaches 0), although Eigen-GCN performs better than the existing GCNs, $\text{MLP}_{\text{feature}}$ reports better results, showing that a graph-based algorithm may not be preferred in those cases after all.
\end{itemize}

\paragraph{Results on Real-world Datasets}\label{sec:nc_setting}
We further experiment on 7 real-world social networks~\cite{traud2012social}. The details of the datasets are provided in Appendix~\ref{sec:additional-details-NC}.
For the base GNN model, we adopt three widely used architectures: GCN by Kipf and Welling~\cite{kipf2017semi}, GAT~\cite{velickovic2018graph}, and GraphSAGE~\cite{hamilton2017inductive}.
We report the results of using GCN in Table~\ref{tab:res1}. The results of using GAT and GraphSAGE show a similar trend and are provided in Appendix~\ref{sec:additional-results-nc} due to the page limit. We also omit the results of $\text{GNN}_{\text{one-hot}}$ since it runs out of memory on most of the datasets. We make the following observations.

    \begin{table}[t]
    \caption{The results of node classification accuracy (\%) on 7 social networks using GCN as the base model. The best results with and without feature information, respectively, are in bold. A, X, Y stands for graph structures, node features, and node labels, respectively.}\label{tab:res1}
    \begin{scriptsize}
    \begin{tabular}{P{0.6cm}| P{1.6cm}  P{1.12cm}  P{1.12cm}  P{1.12cm}  P{1.12cm} P{1.12cm}  P{1.12cm} P{1.12cm} }  \hline
Data &   Method                      & Harvard & Columbia & Stanford & Yale & Cornell & Dartmouth & UPenn \\  \hline
\multirow{4}{*}{A,Y}
&$\text{GCN}_{\text{random}}  $      & $74.6\pm0.5$ & $63.6\pm1.6$ & $68.2\pm1.5$ & $73.6\pm1.2$ & $54.5\pm1.7$ & $73.1\pm1.6$ & $63.0\pm1.2$ \\
&$\text{GCN}_{\text{degree}}  $      & $74.4\pm2.0$ & $63.8\pm2.3$ & $67.8\pm1.6$ & $76.5\pm2.0$ & $56.3\pm1.6$ & $73.3\pm1.4$ & $65.4\pm1.8$ \\ $58.6\pm1.5$
&$\text{GCN}_{\text{DW}}      $      & $82.5\pm1.0$ & $\bf{76.0\pm1.3}$ & $76.6\pm1.3$ & $82.6\pm1.0$ & $71.0\pm2.0$ & $79.3\pm1.7$ & $77.1\pm1.4$ \\
&$\text{Eigen-GCN}_{\text{struc}}$   & $\bf{82.7\pm1.2}$ & $\bf{76.0\pm1.9}$ & $\bf{78.9\pm1.3}$ & $\bf{84.2\pm1.4}$ & $\bf{71.9\pm1.7}$ & $\bf{82.1\pm1.2}$ & $\bf{78.5\pm1.4}$ \\ \hline
\multirow{3}{*}{A,X,Y}
&$\text{GCN}_{\text{feat}}$       & $70.6\pm1.3$ & $74.8\pm1.7$ & $71.3\pm1.6$ & $71.2\pm1.9$ & $67.0\pm1.9$ & $73.1\pm1.6$ & $71.2\pm2.0$ \\
&$\text{GCN}_{\text{feat+DW}}$    & $83.1\pm0.7$ & $77.6\pm1.3$ & $78.3\pm1.4$ & $83.5\pm1.5$ & $73.2\pm2.2$ & $80.8\pm1.3$ & $78.5\pm1.2$ \\
&$\text{Eigen-GCN}_{\text{feat+struc}}$
                                    & $\bf{84.6\pm1.4}$ & $\bf{78.6\pm1.1}$ &$\bf{79.7\pm1.2}$ & $\bf{85.1\pm1.3}$ & $\bf{74.8\pm1.8}$ & $\bf{83.6\pm1.3}$ & $\bf{81.3\pm0.9}$ \\ \hline
    \end{tabular}
    \end{scriptsize}
    \end{table}

\begin{itemize}[leftmargin = 0.6cm]
\item $\text{Eigen-GCN}_{\text{feat+struc}}$ reports the best results on all the datasets and the improvements of $\text{Eigen-GCN}_{\text{feat+struc}}$ compared to $\text{Eigen-GCN}_{\text{feat}}$ are more than 10\% in terms of the classification accuracy on Harvard, Yale, Dartmouth,
and UPenn. The results clearly demonstrate that graph structures are crucial in this task and our proposed method can greatly enhance the existing GCNs in preserving graph structures.
\item When node features are unavailable, $\text{Eigen-GCN}_{\text{struc}}$ also consistently outperforms the other methods. This demonstrates that Eigen-GCN can extract fruitful information from the graph structures and well handle featureless graphs.
\item $\text{GCN}_{\text{DW}}$ and $\text{GCN}_{\text{feat+DW}}$ achieve the second-best results. The heuristic method works reasonably well in node classification task, but is still inferior to Eigen-GCN.
\end{itemize}

We also experiment on four benchmark datasets, namely Cora, Citeseer, Pubmed, and Reddit. We report the results in Appendix~\ref{sec:additional-results-nc}.
The results show that our proposed method achieves comparable performance with the existing GCNs (please see Appendix~\ref{sec:additional-results-nc} for a detailed discussion). The experimental results demonstrate that Eigen-GNN achieves superior performance on structure-driven tasks and does not affect the performance when node features are dominant.

\subsection{Graph Isomorphism Tests}\label{sec:CSLex}

\begin{wrapfigure}{r}{0.5\textwidth}
\vspace{-0.5cm}
\hspace{0.3cm}
        \includegraphics[width=6.3cm]{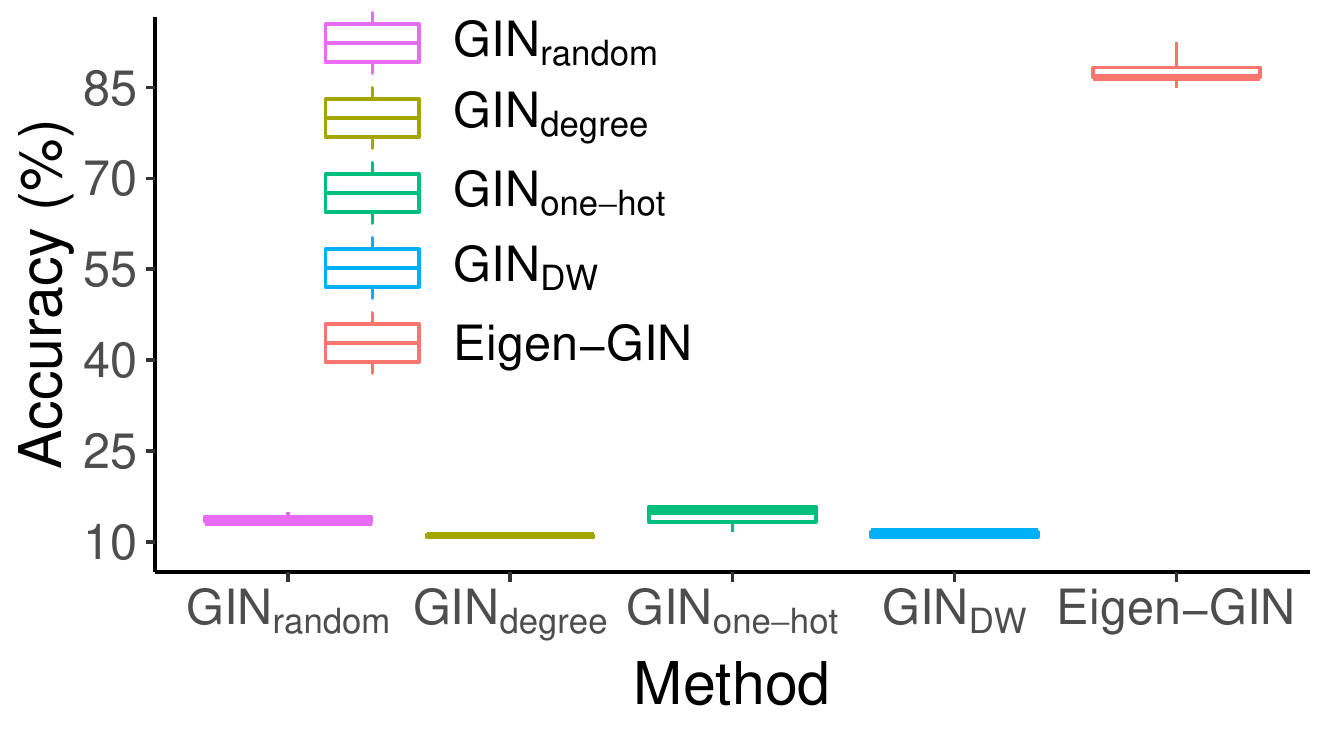}
        \caption{The results of graph isomorphism tests on circulant skip link graphs.}
        \label{fig:iso}
        \vspace{-0.2cm}
\end{wrapfigure}

We further conduct experiments on Circulant Skip Links (CSL) graphs~\cite{murphy19a}, a well-known dataset for graph isomorphism tests, i.e., distinguishing whether two graphs are structurally equivalent. Appendix~\ref{sec:csl} provides more details of CSL graphs and our experimental settings. We adopt GIN~\cite{xu2019powerful} as the baseline GNN model, which is proven to be one of the most powerful message-passing GNN models in graph isomorphism tests\footnote{We do not adopt a more recent approach RP-GIN~\cite{murphy19a} because of its high time complexity.}. Since this dataset does not contain node features, we only report in Fig~\ref{fig:iso} the results of the five methods that do not use node features. We make the following observations.

All the methods except for Eigen-GIN report an accuracy of about 10\%, roughly the same as that of random guessing since the dataset has 10 balanced classes. These results are consistent with the theoretical findings that the original GIN (as well as other message-passing GNNs) cannot distinguish CSL graphs~\cite{murphy19a}. The major reason is that $\text{GIN}_{\text{random}}$, $\text{GIN}_{\text{one-hot}}$, and $\text{GIN}_{\text{DW}}$ do not satisfy permutation-equivariance, a necessary requirement for graph isomorphism tests, and $\text{GIN}_{\text{degree}}$ cannot distinguish graph structures in this case as all nodes have the same degree.

Eigen-GIN reports a remarkably high accuracy. It can recognize CSL graphs well due to two major reasons. First, Eigen-GIN satisfies permutation-equivariance, as proven in Theorem~\ref{thm1}. Second, the eigenspace provides more useful graph structure information than simple heuristics such as degrees.

\paragraph{More experimental results and summary}
 We also conduct experiments on link prediction and analyze the scalability and parameter sensitivity of Eigen-GNN. Due to the page limit, we have to provide the results in Appendix~\ref{sec:LP} and Appendix~\ref{sec:time}, respectively.

In summary, the experimental results show that Eigen-GNN works well with a number of GNN models for different tasks and thus demonstrate its general applicability in enhancing various GNNs in preserving graph structures.

\section{Conclusion}
In this paper, we observe that many GNNs in practice are shallow in nature and do not have a sufficient capability to well preserve graph structure.
We propose Eigen-GNN, a simple yet general and effective plug-in module that enhances the capabilities of GNNs in preserving graph structures. Our extensive experiments demonstrate the effectiveness of Eigen-GNN in a wide spectrum of tasks.
%\clearpage
\section*{Broader Impact}
GNNs can be widely applied to various domains when graph data is available. Some typical examples include social networks, recommendation systems, biological networks, the World Wide Web, and technology networks. Our proposed Eigen-GNN, as a general plug-in to enhance GNNs, could be applied to all these scenarios as long as GNNs are adopted. We expect Eigen-GNN to perform more favorably than the existing GNNs when the tasks are more structure-driven and retain comparable performance when the tasks are feature-driven. Specifically, we find in experiments that social network tasks are more likely to be structure-driven and thus benefit more from our proposed method. As for ethical aspects, we do not foresee that Eigen-GNN should produce any other biased or offensive content than the existing GNNs.

%\begin{ack}
%Use unnumbered first level headings for the acknowledgments. All acknowledgments
%go at the end of the paper before the list of references. Moreover, you are required to declare
%funding (financial activities supporting the submitted work) and competing interests (related financial activities outside the submitted work).
%More information about this disclosure can be found at: \url{https://neurips.cc/Conferences/2020/PaperInformation/FundingDisclosure}.
%
%Do {\bf not} include this section in the anonymized submission, only in the final paper. You can use the \texttt{ack} environment provided in the style file to autmoatically hide this %section in the anonymized submission.
%\end{ack}

\bibliographystyle{plain}
\bibliography{cite}
\clearpage
\appendix
    \begin{table}
    \centering
    \caption{GNN methods following Eq.~\eqref{eq:GCN1} and their corresponding graph structure functions. $\mathbf{A}_P$ is the Positive Pointwise Mutual Information (PPMI) matrix~\cite{zhuang2018dual}.}\label{tab:framework}
    \begin{tabular}{c | c} \hline
        Method &  $\mathcal{F}\left( \mathbf{A} \right)$ \\ \hline
        GCN~\cite{kipf2017semi}, SGC~\cite{wu2019simplifying}        & $\tilde{\mathbf{D}}^{-\frac{1}{2}} \tilde{\mathbf{A}} \tilde{\mathbf{D}}^{-\frac{1}{2}}$ \\
        DCNN~\cite{atwood2016diffusion} & $\left( \mathbf{D}^{-1} \mathbf{A} \right)^K$ \\
        DGCN~\cite{zhuang2018dual}      & $\mathbf{D}^{-\frac{1}{2}}_P \mathbf{A}_P \mathbf{D}^{-\frac{1}{2}}_P$ \\
        PPNP~\cite{klicpera2019predict} & $\alpha\left( \mathbf{I}_N - \left( 1 - \alpha \right) \tilde{\mathbf{D}}^{-\frac{1}{2}} \tilde{\mathbf{A}} \tilde{\mathbf{D}}^{-\frac{1}{2}} \right)^{-1}$ \\ \hline
    \end{tabular}
    \end{table}
\section{Additional Experimental Results}\label{sec:additional-results}
\subsection{Node Classification}\label{sec:additional-results-nc}
The results of node classification on 7 real-world social networks using GAT and GraphSAGE as the base GNN model are reported in Table~\ref{tab:nc_add1} and Table~\ref{tab:nc_add2}, respectively.
In general, the results show similar trends as using GCN, i.e., Table~\ref{tab:res1} in Section~\ref{sec:nc_setting}. One exception in Table~\ref{tab:nc_add2} using GraphSAGE as the base GNN is that $\text{SAGE}_{\text{DW}}$ slightly outperforms $\text{Eigen-SAGE}_{\text{struc}}$ on two datasets (Columbia and Cornell). However, when combining with features, $\text{Eigen-SAGE}_{\text{feat+struc}}$ still outperforms $\text{SAGE}_{\text{feat+DW}}$ on these two datasets.

    \begin{table}[ht]
    \caption{The results of node classification accuracy (\%) on 7 social networks using GAT as the base model. The best results with and without feature information, respectively, are in bold. A, X, Y stands for graph structures, node features, and node labels, respectively.}\label{tab:nc_add1}
    \begin{scriptsize}
    \begin{tabular}{P{0.6cm}| P{1.6cm}  P{1.12cm}  P{1.12cm}  P{1.12cm}  P{1.12cm} P{1.12cm}  P{1.12cm} P{1.12cm} }  \hline
Data &   Method                      & Harvard & Columbia & Stanford & Yale & Cornell & Dartmouth & UPenn \\  \hline
\multirow{4}{*}{A,Y}
&$\text{GAT}_{\text{random}}  $      & $\bf{81.5\pm0.7}$ & $74.4\pm1.2$ & $74.8\pm1.6$ & $80.7\pm1.8$ & $67.4\pm2.5$ & $79.6\pm1.2$ & $74.8\pm1.5$ \\
&$\text{GAT}_{\text{degree}}  $      & $67.5\pm6.1$ & $63.7\pm4.1$ & $63.2\pm2.8$ & $70.5\pm3.5$ & $54.2\pm2.6$ & $68.5\pm3.2$ & $60.9\pm3.3$ \\
&$\text{GAT}_{\text{DW}}      $      & $\bf{81.5\pm1.3}$ & $75.4\pm1.7$ & $76.3\pm2.2$ & $81.2\pm2.3$ & $70.1\pm2.9$ & $77.4\pm2.2$ & $75.1\pm1.9$ \\
&$\text{Eigen-GAT}_{\text{struc}}$   & $\bf{81.5\pm1.8}$ & $\bf{77.1\pm1.4}$ & $\bf{78.1\pm1.5}$ & $\bf{83.9\pm1.3}$ & $\bf{72.0\pm2.0}$ & $\bf{81.5\pm2.0}$ & $\bf{78.5\pm1.1}$ \\ \hline
\multirow{3}{*}{A,X,Y}
&$\text{GAT}_{\text{feat}}$       & $73.7\pm1.6$ & $73.9\pm2.2$ & $71.5\pm2.0$ & $72.7\pm3.5$ & $63.9\pm2.2$ & $74.9\pm2.1$ & $69.0\pm3.5$ \\
&$\text{GAT}_{\text{feat+DW}}$    & $84.3\pm0.6$ & $78.4\pm1.7$ & $76.0\pm3.4$ & $80.9\pm2.0$ & $73.3\pm2.6$ & $78.1\pm1.7$ & $75.5\pm2.2$ \\
&$\text{Eigen-GAT}_{\text{feat+struc}}$
                                    & $\bf{85.5\pm1.3}$ & $\bf{78.9\pm1.1}$ & $\bf{79.7\pm1.4}$ & $\bf{84.3\pm1.8}$ & $\bf{73.6\pm1.8}$ & $\bf{82.7\pm1.2}$ & $\bf{80.0\pm1.7}$ \\ \hline
    \end{tabular}
    \end{scriptsize}
    \end{table}

    \begin{table}[ht]
    \caption{The results of node classification accuracy (\%) on 7 social networks using GraphSAGE as the base model. The best results with and without feature information, respectively, are in bold. A, X, Y stands for graph structures, node features, and node labels, respectively.}\label{tab:nc_add2}
    \begin{scriptsize}
    \begin{tabular}{P{0.6cm}| P{1.6cm}  P{1.12cm}  P{1.12cm}  P{1.12cm}  P{1.12cm} P{1.12cm}  P{1.12cm} P{1.12cm}}  \hline
Data &   Method                      & Harvard & Columbia & Stanford & Yale & Cornell & Dartmouth & UPenn \\  \hline
\multirow{4}{*}{A,Y}
&$\text{SAGE}_{\text{random}}  $     & $14.2\pm1.2$ & $18.8\pm1.4$ & $17.8\pm0.7$ & $17.0\pm1.9$ & $18.3\pm1.1$ & $17.8\pm1.0$ & $18.4\pm1.2$ \\
&$\text{SAGE}_{\text{degree}}  $     & $25.9\pm2.3$ & $25.9\pm2.3$ & $27.1\pm2.0$ & $30.0\pm3.0$ & $34.0\pm1.9$ & $25.7\pm2.3$ & $24.1\pm1.7$ \\ $25.9\pm1.5$
&$\text{SAGE}_{\text{DW}}      $     & $82.6\pm1.0$ & $\bf{77.5\pm1.3}$ & $76.0\pm1.5$ & $82.2\pm1.1$ & $\bf{71.5\pm1.8}$ & $80.5\pm1.5$ & $77.6\pm1.1$ \\
&$\text{Eigen-SAGE}_{\text{struc}}$  & $\bf{83.1\pm0.9}$ & $76.4\pm1.6$ & $\bf{78.2\pm1.0}$ & $\bf{84.9\pm1.3}$ & $70.7\pm1.4$ & $\bf{82.4\pm1.6}$ & $\bf{78.3\pm1.5}$ \\ \hline
\multirow{3}{*}{A,X,Y}
&$\text{SAGE}_{\text{feat}}$      & $76.6\pm1.5$ & $77.6\pm1.5$ & $73.3\pm1.5$ & $79.2\pm2.5$ & $62.7\pm1.9$ & $78.5\pm1.1$ & $71.4\pm1.7$ \\
&$\text{SAGE}_{\text{feat+DW}}$   & $80.4\pm0.9$ & $77.5\pm2.0$ & $75.1\pm1.6$ & $82.6\pm1.1$ & $71.2\pm1.7$ & $77.3\pm1.3$ & $77.4\pm1.1$ \\
&$\text{Eigen-SAGE}_{\text{feat+struc}}$
                                    & $\bf{84.8\pm1.2}$ & $\bf{80.0\pm1.3}$ & $\bf{78.0\pm0.9}$ & $\bf{86.2\pm1.4}$ & $\bf{73.1\pm2.0}$ & $\bf{84.6\pm1.4}$ & $\bf{79.5\pm1.7}$ \\ \hline
    \end{tabular}
    \end{scriptsize}
    \end{table}

We also experiment on four benchmark datasets commonly used in GNNs: Cora, Citeseer, Pubmed, and Reddit. The results are shown in Table~\ref{tab:nc_add3}. For simplicity, we only adopt GCN by Kipf and Welling~\cite{kipf2017semi} as the base GNN model. We make the following observations.
\begin{itemize}[leftmargin = 0.5cm]

\item Similar to Section~\ref{sec:nc_setting}, when no node feature is available, $\text{Eigen-GCN}_{\text{struc}}$ reports the best results, demonstrating that Eigen-GCN can better preserve graph structures.
\item When features are available, $\text{GCN}_{\text{feat}}$ performs the best on three citation graphs and highly competently on Reddit, showing that features are dominant on these benchmark datasets. The results are consistent with the literature~\cite{wu2019simplifying,maehara2019revisiting}, which show that features contain the ``true signals'' for those node classification tasks.
\item $\text{Eigen-GCN}_{\text{feat+struc}}$ has comparable performance with $\text{GCN}_{\text{feat}}$ on the three citations graphs and is even better than $\text{GCN}_{\text{feat}}$ on Reddit. These results show that expanding the initial bases with the eigenspace does not impair GNNs in feature-driven tasks. Thus, Eigen-GNN can be adopted as a default module if we are not sure whether a task is feature-driven or structure-driven.
\end{itemize}

\begin{table}[!ht]
    \centering
    \caption{The results of node classification accuracy (\%) on the benchmark datasets. The best results with and without feature information, respectively, are in bold. A, X, Y stands for graph structures, node features, and node labels, respectively.}\label{tab:nc_add3}
    \begin{tabular}{ c | c  c  c  c  c}               \hline
     Data & Method    &  Cora        &  Citeseer    &    Pubmed   & Reddit   \\  \hline
       \multirow{5}{*}{A,Y}
                    & $\text{GCN}_{\text{random}}  $  & $23.5\pm1.6$ & $21.2\pm1.1$ & $32.6\pm1.0$  & $86.1\pm0.3$ \\
                    & $\text{GCN}_{\text{degree}}  $  & $33.5\pm2.4$ & $30.2\pm0.9$ & $34.9\pm1.3$  & $83.0\pm0.4$ \\
                    & $\text{GCN}_{\text{one-hot}} $  & $66.3\pm0.6$ & $45.2\pm1.1$ & $64.3\pm0.9$  & Out of memory \\
                    & $\text{GCN}_{\text{DW}}      $  & $70.6\pm1.2$ & $47.7\pm1.1$ & $69.3\pm1.2$  & $\bf{94.3\pm0.1}$ \\
                    & $\text{Eigen-GCN}_{\text{struc}}$  & $\bf{71.0\pm0.5}$ & $\bf{49.3\pm0.6}$ & $\bf{73.8\pm0.3}$ & $\bf{94.3\pm0.0}$ \\ \hline
       \multirow{3}{*}{A,X,Y}
                    & $\text{GCN}_{\text{feat}}$  & $\bf{81.5\pm0.4}$ & $\bf{70.6\pm0.8}$ & $\bf{78.6\pm0.4}$   & $96.4\pm0.0$ \\
                    & $\text{GCN}_{\text{feat+DW}}$  & $76.8\pm0.5$ & $61.8\pm0.6$ & $76.3\pm0.5$ & $\bf{96.6\pm0.1}$ \\
                    & $\text{Eigen-GCN}_{\text{feat+struc}}$  & $78.9\pm0.7$ & $66.5\pm0.3$ & $\bf{78.6\pm0.1}$ & $\bf{96.6\pm0.1}$ \\ \hline
    \end{tabular}
    \end{table}

\subsection{Link Prediction}\label{sec:LP}

Link prediction is to predict which pairs of nodes in a graph are most likely to form edges. We adopt eight benchmark datasets from~\cite{NIPS2018_7763} with the same experimental setting.
The details of the datasets and experimental settings are provided in Appendix~\ref{sec:additional-details-LP}.
We use SEAL~\cite{NIPS2018_7763} as the baseline GNN model, a state-of-the-art GNN architecture specifically designed for link prediction. The architecture is kept the same as in the original paper.

The results are reported in Table~\ref{tab:LP}. We exclude the five baselines that do not use node features since SEAL has specifically designed those features and cannot function without them. We make the following observations.

 \begin{table}[h]
    \caption{The average precision of link prediction (\%). The best results are highlighted in bold. }\label{tab:LP}
    \begin{scriptsize}
    \begin{tabular}{ c | P{1.1cm}  P{1.1cm}  P{1.1cm}  P{1.1cm}  P{1.1cm}  P{1.1cm}  P{1.1cm}  P{1.1cm} }               \hline
           Dataset & C.elegans & E.coli & NS      &PB      &Power   & Router  &USAir   &Yeast    \\ \hline
    $\text{SEAL}   $ & $77.6\pm0.9$ & $91.5 \pm 0.8 $ & $96.8 \pm 1.8 $ &$87.3 \pm 0.3 $ &$69.9 \pm 1.6 $ &$88.2 \pm 1.0 $ &$\bf{90.3 \pm 1.6} $ &$\bf{93.7 \pm 0.3} $ \\
    $\text{SEAL}_{\text{DW}}$ & $77.1\pm1.5$ & $92.1 \pm 0.7 $ & $97.0 \pm 1.2 $ &$87.5 \pm 0.2 $ &$69.8 \pm 1.5 $ &$ 87.9 \pm 1.3   $ &   $ 89.9 \pm 1.9 $ &$ 93.6 \pm  0.5 $     \\
    $\text{Eigen-SEAL}$ &$  \bf{79.5 \pm 0.8}^* $ &$  \bf{92.5 \pm  0.6}^* $ &$  \bf{97.2 \pm  0.6} $ & $  \bf{87.8 \pm  0.4}^* $ &$  \bf{73.2 \pm  2.4}^* $ &$  \bf{88.3 \pm  1.2} $ &$  \bf{90.3 \pm  1.2} $ & $  93.6 \pm  0.4 $ \\ \hline
     Gain$\dagger$ & $  +1.9 $ & $  +0.4 $ & $  +0.2 $ & $  +0.3 $ & $  +3.3 $ & $  +0.1 $ & $  0.0$ & $  -0.1$ \\  \hline
    \end{tabular}
    \\ $\dagger$: Gain is the relative improvement of Eigen-SEAL compared to the better of the other two methods.
    \\ *: The improvement of bolded results over non-bolded results is statistically significant at 0.05-level paired t-test.
    \end{scriptsize}
    \end{table}

\begin{itemize}[leftmargin = 0.6cm]
\item Eigen-SEAL reports significantly better results than the two baselines on four out of the eight datasets (with the rest four datasets showing no significant differences). This indicates that graph structures are important in link prediction tasks. The finding is consistent with the literature~\cite{liben2007link}.
\item Although SEAL is specifically designed for link prediction and Eigen-GNN does not target at any specific task, Eigen-SEAL reports better performance.  This demonstrates the general effectiveness of Eigen-GNN.
\item $\text{SEAL}_{\text{feat+DW}}$ fails to outperform SEAL on any dataset\footnote{Though $\text{SEAL}_{\text{feat+DW}}$ seems to improve SEAL on E.coli, NS, and PB, the improvement is not statistically significant under 0.05 paired t-test.}. This shows that DeepWalk cannot enhance SEAL in link prediction. The results are consistent with the original paper~\cite{NIPS2018_7763}.
\end{itemize}

\subsection{Scalability and Parameter Sensitivity}\label{sec:time}
\paragraph{Scalability} Since Eigen-GNN conducts the same calculation as base GNN models in all the hidden layers, we only report the runtime of calculating the eigenbasis, which is the extra cost caused by Eigen-GNN. Specifically, we generate random graphs of different sizes using the Erdos Renyi model~\cite{erdHos1960evolution}. Figure~\ref{fig:scala} shows the runtime when fixing either the number of nodes to ten thousand or fixing the number of edges to one million, and varying the other factor. The time of calculating the eigenbasis increases roughly linearly with respect to the number of nodes and the number of edges in graphs. In addition, even for a large graph with ten thousand nodes and five million edges, the running time is no more than a few seconds on a single PC. The results show that Eigen-GNN is scalable to large-scale graphs.

\begin{figure}[t]
\begin{subfigure}{.54\textwidth}
  \centering
  \includegraphics[height=3cm]{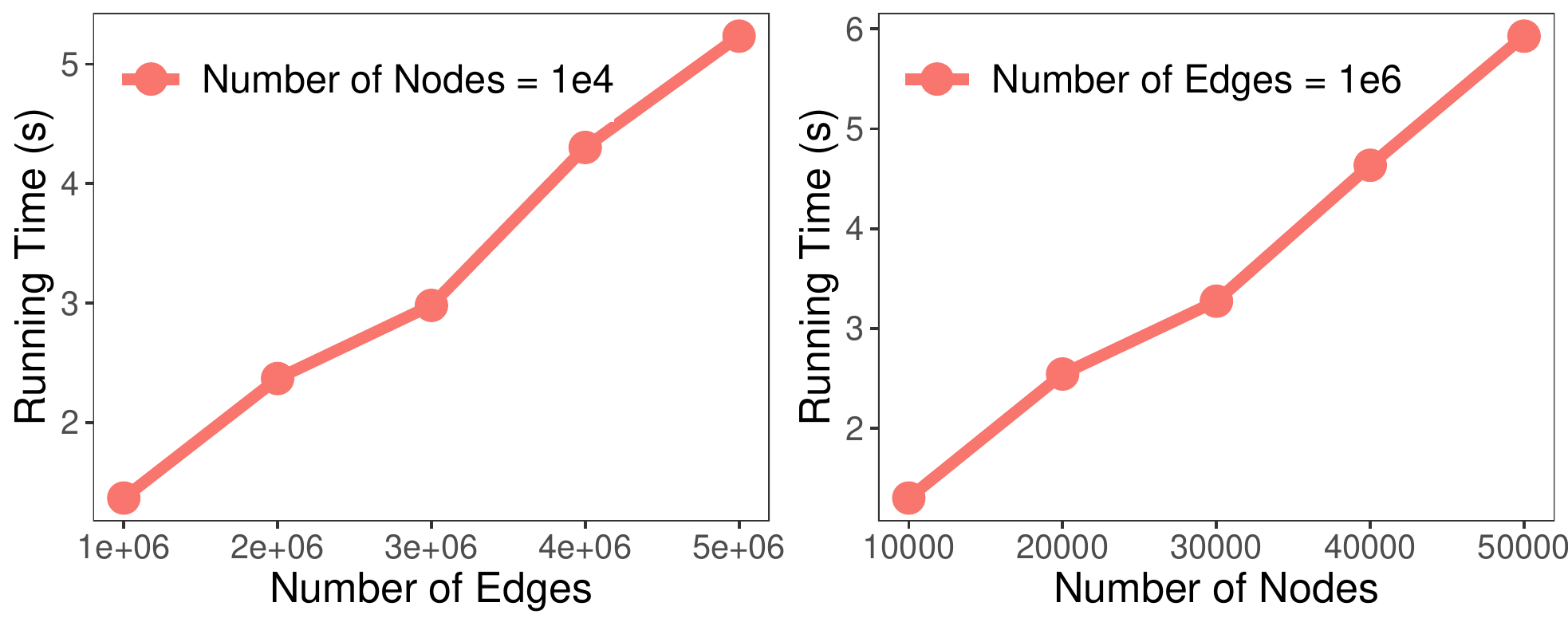}\\
  \caption{}
  \label{fig:scala}
\end{subfigure}
\begin{subfigure}{.46\textwidth}
  \centering
  \includegraphics[height=3cm]{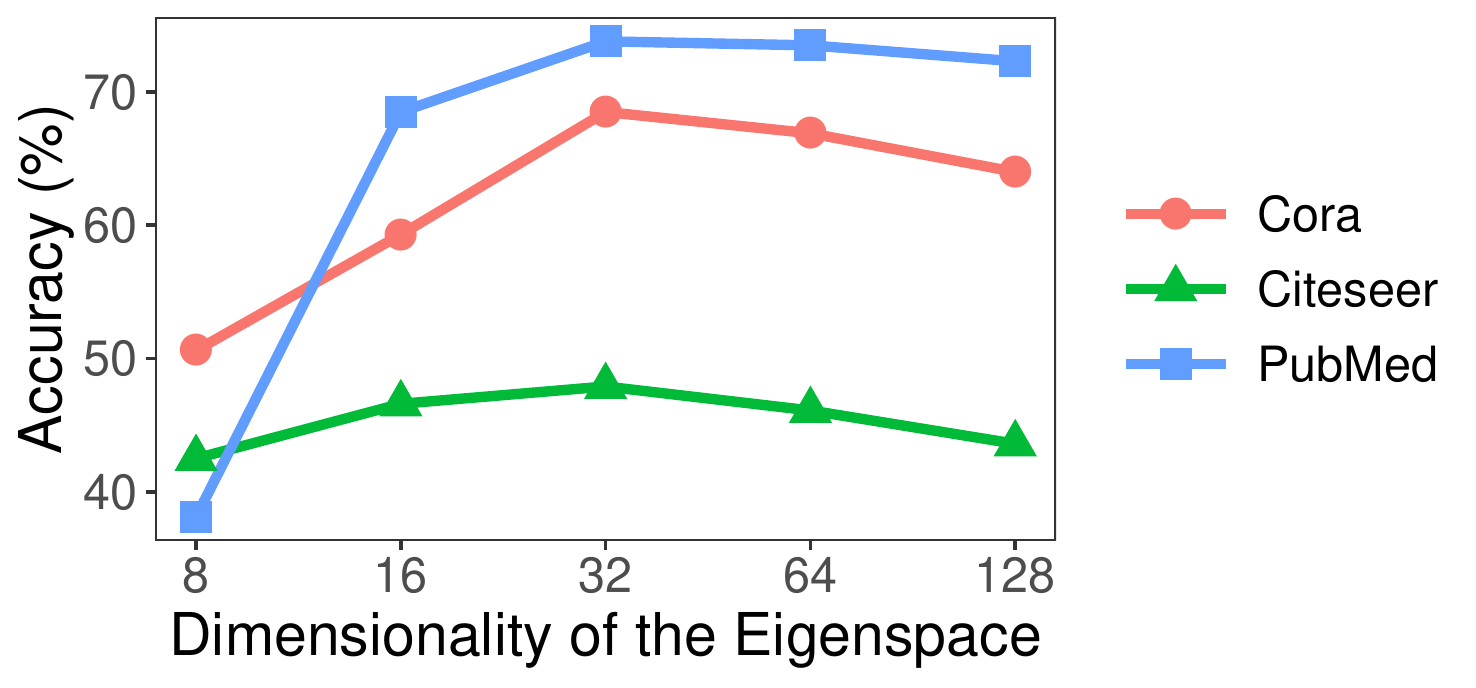}\\
  \caption{}
  \label{fig:analysis}
\end{subfigure}
\caption{Scalability and parameter sensitivity. (a) The running time of calculating the eigenbasis grows linearly with respect to the number of nodes and the number of edges in the graph, respectively. (b) The node classification accuracy of Eigen-GNN with different eigenspace dimensionality.}
\end{figure}

\paragraph{Parameter Sensitivity} Eigen-GNN has only one parameter, the dimensionality $d$ of the eigenspace. To test the parameter sensitivity, we follow the same experimental setting as in Appendix~\ref{sec:additional-results-nc} by adopting GCN~\cite{kipf2017semi} as the base GNN model and vary $d$ in $\left\{8,16,32,64,128\right\}$. Figure~\ref{fig:analysis} shows the node classification results on the three citations graphs without using node features. The results of the other tasks on the corresponding datasets share similar patterns. When the dimensionality $d$ increases, the accuracy of the model increases at first but tends to saturate or even decreases if $d$ becomes too large. A plausible reason is that, if the dimensionality of the eigenspace is too small, the model does not have enough capacities to learn useful graph structures. If the eigenspace grows too large, noises are likely to be introduced into the model.

\section{Additional Experimental Details for Reproducibility}\label{sec:additional-details}
\subsection{Synthetic datasets generation details}\label{sec:additional-details-syn}
As introduced in Section~\ref{sec:preli}, we use the Stochastic Blockmodel to generate the graph structures. Specifically, we generate $5,000$ nodes in a graph, which are randomly assigned to one of the ten different communities. Within each community, nodes have a probability $prob_{in} = 0.025$ to form edges, while nodes in different communities have a probability $prob_{out} = 0.001$ to form edges.

For node features, we also divide nodes into ten groups. For each group, we generate a group vector following $\mathcal{N}(0,4\mathbf{I})$ with dimensionality $32$. We constrain that the minimum Euclidean distance between every two group vectors is larger than $1$. Then, node features are randomly generated following i.i.d standard normal distribution $\mathcal{N}(0,\mathbf{I})$ around the group vectors of the groups that the nodes belong to.

In summary, each synthetic dataset has $5,000$ nodes, $85,294$ edges, $32$ features, and $10$ labels/classes.  Here, a class contains all nodes carrying the same label. The differences among different datasets are the nodes carrying different labels. In Figure~\ref{fig:pre}, the number of randomly selected nodes per class used for training is used as the x-axis and we randomly selected another $200$ nodes per class for validation. The rest nodes are used for testing. In Figure~\ref{fig:pre2}, we use $50$ nodes per class for training, $150$ nodes per class for validation, and the rest for testing.

\subsection{Node classification datasets details}\label{sec:additional-details-NC}
\begin{itemize}[leftmargin = 0.6cm]
\item Harvard, Columbia, Stanford, Yale, Cornell, Dartmouth, and UPenn~\cite{traud2012social}\footnote{\url{https://archive.org/details/oxford-2005-facebook-matrix}}:
these are Facebook social networks for different colleges/universities. Edges represent intra-school links of users and node attributes correspond to user profiles such as gender, major, dorm/house, etc. We use the class year as ground-truth labels. We preprocess the datasets by using a one-hot encoding of categorical node features and removing node features/labels which occur less than 0.1\%/1\% among all the nodes.
\item Cora\footnote{\label{note1}\url{https://github.com/tkipf/gcn}}, Citeseer\textsuperscript{\ref{note1}}, Pubmed\textsuperscript{\ref{note1}}~\cite{sen2008collective}: three citation graphs where nodes represent papers and edges represent citations between papers. The datasets also contain bag-of-words features and ground-truth topics as labels of the papers.
\item Reddit\footnote{\label{note2}\url{http://snap.stanford.edu/graphsage/}}~\cite{hamilton2017inductive}: an online discussion forum for users where nodes are posts and two nodes are connected if they are commented by the same user. Each post also contains a low-dimensional word vector as features. The task is to predict which community the posts belong to.
\end{itemize}

The statistics of the datasets are summarized in Table~\ref{tab:stat1}. For the Facebook social networks, we use $20$ nodes per class for training, $30$ nodes per class for validation, and the rest for testing. For the other four benchmark datasets, we adopt the fixed training/validation/testing split that came with the datasets. Similar results are observed in random splits.
    \begin{table}[ht]
    \caption{Statistics of the datasets used for node classification.}\label{tab:stat1}
    \centering
    \begin{tabular}{c | c  c  c  c c} \hline
    Dataset  & Type     & \# Nodes   & \#Edges  & \#Classes & \#Features \\ \hline
    Harvard  & Social & 15,126 & 1,649,234 &10 & 136 \\
    Columbia & Social & 11,770 & 888,666   & 7 & 197 \\
    Stanford & Social & 11,621 & 1,136,660 & 8 & 225 \\
    Yale     & Social & 8,578  & 810,900   & 8 & 146 \\
    Cornell  & Social & 18,660 & 1,581,554 & 7 & 253 \\
    Dartmouth& Social & 7,694  & 608,152   & 9 & 178 \\
    UPenn    & Social & 14,916 & 1,373,002 & 7 & 204 \\
    Cora     & Citation & 2,708   & 5,429  & 7       & 1,433    \\
    Citeseer & Citation & 3,327   & 4,732  & 6       & 3,703    \\
    Pubmed   & Citation & 19,717  & 44,338 & 3       & 500      \\
    Reddit   & Social   & 232,965 & 11,606,919 & 41  & 602      \\ \hline
    \end{tabular}
    \end{table}
\subsection{Circulant skip link graphs details}\label{sec:csl}
A basic Circulant Skip Link (CSL) graph $\pmb{G}_{N,R}$ is an undirected graph, where $\left\{1,...,N\right\}$ is the set of $N$ nodes and the edges consist of a cycle and a set of skip links. Denote $\mathbf{A}$ as the adjacency matrix. The cycle is formulated as:
\begin{gather}
    \mathbf{A}_{j,j+1} = \mathbf{A}_{j+1,j} = 1, \forall 1 \leq j < N \\
    \mathbf{A}_{1,N} = \mathbf{A}_{N,1} = 1.
\end{gather}
The skip links, controlled by an interval parameter $R$ satisfying $1 < R < N$, are defined as:
\begin{gather}
    \mathbf{A}_{i,j} = \mathbf{A}_{j,i} = 1, \text{if } \left|j-i\right| = R\text{ or }N-R \text{ mod } N, \forall 1 \leq i,j \leq N.
\end{gather}

Figure~\ref{fig:csl} shows examples of $\pmb{G}_{13,2}$ and $\pmb{G}_{13,3}$, i.e., two CSL graphs with $13$ nodes and with skip links of intervals $2$ and $3$, respectively.
Intuitively, basic CSL graphs $\pmb{G}_{N,R}$ are 4-regular graphs (i.e., the degree of all nodes is 4) by connecting every ``adjacent'' node pair and every node pair that is ``R-hops'' away.
The full CSL graph set includes the basic CSL graphs $\pmb{G}_{N,R}$ and all their permutations, i.e.,
\begin{equation}
    \pmb{G}_{N} = \left\{ \mathcal{S}_N\left( \pmb{G}_{N,R} \right), \forall 1 < R < N,\forall \mathcal{S}_N \right\},
\end{equation}
where $\mathcal{S}_N(\cdot)$ is any permutation of $N$ node IDs.

\begin{figure}[ht]
  \centering
  \includegraphics[width=6cm]{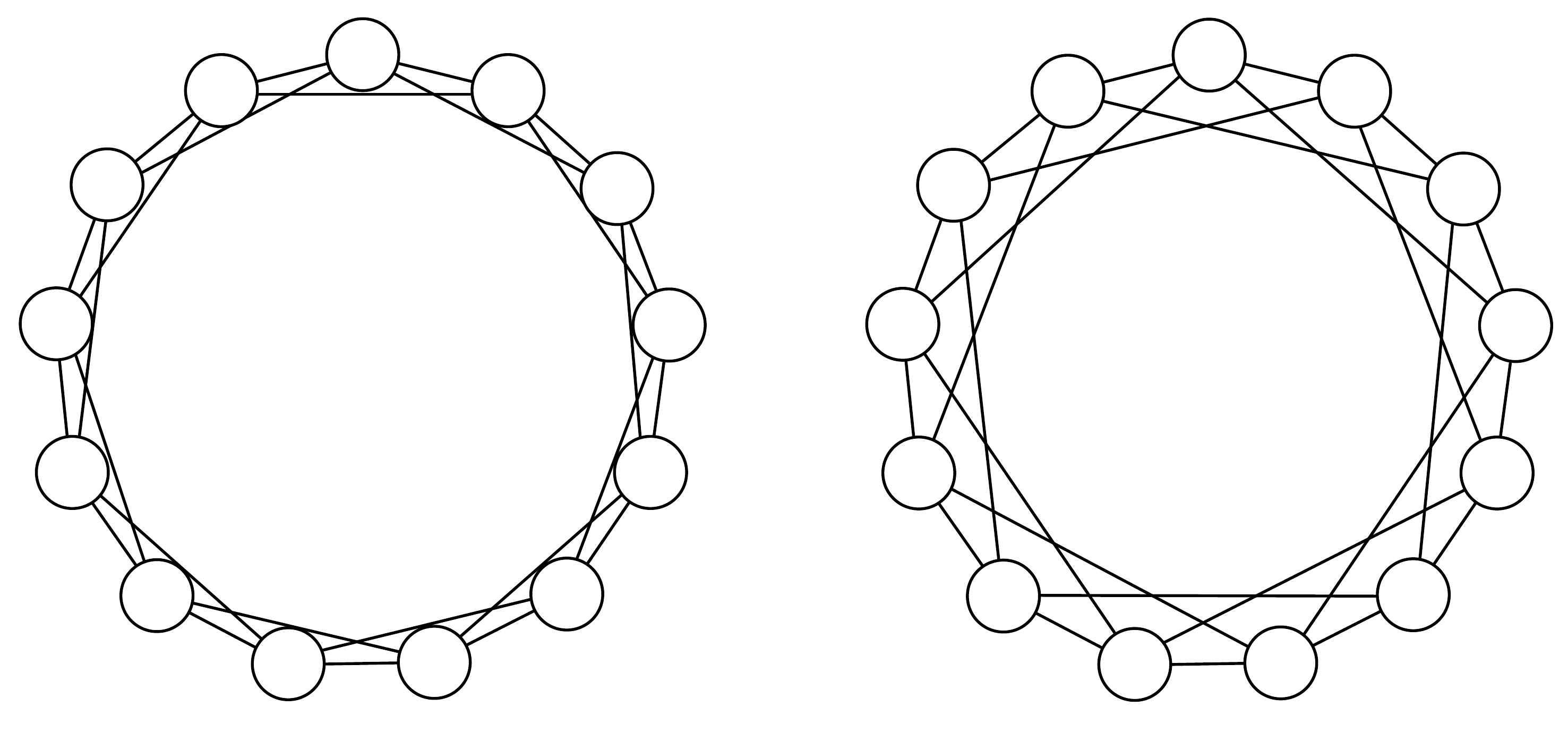}\\
  \caption{An example of CSL graphs $\pmb{G}_{13,2}$ and $\pmb{G}_{13,3}$. Though these two graphs are non-isomorphism, their structures are extremely similar that the existing GNNs fail to distinguish them. The image is adapted from \cite{murphy19a}.}\label{fig:csl}
\end{figure}

CSL graphs are widely adopted for graph isomorphism tests since their structures are highly regular and similar, and all nodes have the same degree. For example, it is known that $\pmb{G}_{41}$ is composed of 10 isomorphism classes: $$\pmb{G}_{41} = \left\{\mathcal{S}_{41} \left(\pmb{G}_{41,R} \right)|R\in\left\{2, 3, 4, 5, 6, 9, 11, 12, 13, 16\right\}\right\}.$$ Although there exist known mathematical approaches to solve graph isomorphism tests for CSL graphs~\cite{muzychuk2004solution}, it still poses great challenges for machine learning models, including the existing GNNs, to distinguish them if no prior knowledge is used~\cite{murphy19a,chen2019equivalence}.

Specifically, following the experimental setting in~\cite{murphy19a}, we consider the aforementioned CSL graphs with 41 nodes and 10 isomorphism classes. Using the isomorphism classes as labels for graphs, the graph isomorphism test can be transformed into a graph classification problem. For each isomorphism class, i.e., a graph label, we randomly generate 60
isomorphic CSL graphs belonging to that class. As a result, the dataset contains 600 graphs with 10 balanced classes. Following~\cite{murphy19a}, we adopt a 5-fold cross-validation.

\subsection{Link prediction datasets details}\label{sec:additional-details-LP}
We adopt the following eight link prediction datasets\footnote{\url{https://github.com/muhanzhang/SEAL}}:
\begin{itemize}[leftmargin = 0.6cm]
\item C.elegans: the neural network of the worm C.elegans. %
\item E.coli: a pairwise reaction network of metabolites in E.coli.
\item NS: a collaboration network between researchers, where nodes represent authors and edges correspond to co-authorships. %
\item PB: a graph formed by US political blogs where edges represent hyperlinks between blogs. %
\item Power: an electrical grid of the western US, where edges represent high-voltage transmission lines between facilities. %
\item Router: a router-level Internet connection graph. %
\item USAir: a graph from US Airlines with nodes representing airports and edges representing airlines. %
\item Yeast: a protein-protein interaction network in yeast. %
\end{itemize}
    \begin{table}[h]
    \caption{Statistics of the datasets used for link prediction.}    \label{tab:stat2}
    \centering
    \begin{tabular}{c | c  c  c c} \hline
    Dataset   & Type & \# Nodes & \# Edges & Degree \\ \hline
    C.elegans & Biology       & 297   & 4,296  & 14.5 \\
    Ecoli     & Biology       & 1,805 & 29,320 & 16.2 \\
    NS        & Collaboration & 1,589 & 5,484  & 3.5  \\
    PB        & Social        & 1,222 & 33,428 & 27.4 \\
    Power     & Industry      & 4,941 & 13,188 & 2.7  \\
    Router    & Internet      & 5,022 & 12,516 & 2.5  \\
    USAir     & Transportation& 332  & 4,252  & 12.8 \\
    Yeast     & Biology       & 2,375 & 23,386 & 9.9  \\ \hline
    \end{tabular}
    \end{table}
The statistics of the datasets are summarized in Table~\ref{tab:stat2}. Following~\cite{NIPS2018_7763}, we randomly split the edges of the graph into 50\%-20\%-30\% parts, and use them for training, validation, and testing, respectively. In splitting the datasets, we maintain that each node has at least one edge in the training set. The same number of edges are sampled from the non-existing links (i.e., node pairs that do not have edges) as negative samples.

\subsection{Source codes adopted} We use the following publicly available implementations of base GNN models and DeepWalk:
\begin{itemize}[leftmargin = 0.6cm]
\item GCN~\cite{kipf2017semi}: \url{https://github.com/tkipf/gcn}
\item StochasticGCN~\cite{chen2018stochastic}: \url{https://github.com/thu-ml/stochastic_gcn}
\item GIN~\cite{xu2019powerful}: \url{https://github.com/PurdueMINDS/RelationalPooling}
\item SEAL~\cite{NIPS2018_7763}: \url{https://github.com/muhanzhang/SEAL}
\item DeepWalk~\cite{perozzi2014deepwalk}: \url{https://github.com/phanein/deepwalk}
\item GAT~\cite{velickovic2018graph} and GraphSAGE~\cite{hamilton2017inductive}: \url{https://github.com/rusty1s/pytorch_geometric}
\end{itemize}

\subsection{Hyper-parameters}
We adopt the following hyper-parameters.
\begin{itemize}[leftmargin = 0.6cm]
\item Synthetic: the number of graph convolution layers and fully connected layers are given in Section~\ref{sec:preli1}, with each layer containing 16 units. The dropout rate is searched from $\left\{0,0.5\right\}$ and L2 regularization is $5\times 10^{-4}$. The learning rate is searched from $\left\{0.005,0.01,0.025\right\}$. The maximum number of epochs is 400 with an early stopping round 100. We also test whether adding residual connections can improve the performance of GCNs and add residual connections in GCN2, GCN3, and GCN5. The hyper-parameters for DeepWalk are the default settings in the implementations of the authors, i.e., the number of walks 10, the walk length 40, the window size 5, and the dimensionality is set to 32 to fairly compare with GCNs with features as input. The dimensionality of the eigenspace is also set to $d=32$ and we set $f(x) = x$.
\item Harvard, Columbia, Stanford, Yale, Cornell, Dartmouth, and UPenn: we set the same hyper-parameters for GCN, GAT, and GraphSAGE, which is a two-layer architecture with the hidden layer containing 64 units (for GAT, 8 heads with each head containing 8 units), the learning rate is 0.01 with weight decay $5\times10^{-4}$, the maximum number of epochs is 400, and we select the epoch with the highest validation accuracy. The dropout is searched from $\left\{0, 0.5\right\}$ and the dimensionality of the eigenspace is searched from $\left\{128,256 \right\}$. We set $f(x)$ as a normalization function according to the Frobenius norm, i.e., $f(\mathbf{Z}) = \frac{\mathbf{Z}}{\left\| \mathbf{Z}\right\|_F}$.
\item Cora, Citeseer, and Pubmed: we use the same GCN structure and hyper-parameters as in~\cite{kipf2017semi}, i.e., a two-layer GCN with the hidden layer containing 16 units, the dropout rate 0.5, L2 regularization $5\times10^{-4}$, the learning rate 0.01. The maximum number of epochs is 400 with an early stopping round 100. The dimensionality of the eigenspace is set to $d=32$ and $f(x)$ is the aforementioned normalization function according to the Frobenius norm.
\item Reddit: since the dataset has more than 200 thousand nodes and 11 million edges, a full-batch training of GNN is infeasible. Instead, we adopt the neighborhood sampling strategy proposed in Stochastic GCN~\cite{chen2018stochastic} for all the methods. We use the exact same hyper-parameters and sampling strategy suggested in~\cite{chen2018stochastic}, i.e., the GraphSAGE mean pooling architecture, two graph convolution layers with the hidden layer size of 128 units, sampling two neighbors per node, no weight decay, the dropout rate 0.2, with layer-normalization, the batch-size 512, and a maximum of 30 epochs. The dimensionality of the eigenspace is set to $d=128$ and we set $f(x)$ as the aforementioned normalization based on Frobenius norm.
\item CSL: we tested using the default GIN structure (5 GNN layers, 2 MLP layers)~\cite{xu2019powerful} and a simplified GIN structure (1 GNN layer, 1 MLP layer). Since no substantial difference is observed, we adopt the latter due to its simplicity. Other settings are the same as in~\cite{murphy19a}, i.e., no dropout or batch normalization, all layers having 16 hidden units, training epsilon via back-propagation, the learning rate 0.01, a maximum of 200 epochs, and no early stopping. The dimensionality of the eigenspace is $d=16$ and we set $f(x) = \left| x\right|$ to ensure permutation-equivariance.
\item The eight link prediction datasets: we use the default SEAL architecture~\cite{chen2018stochastic}, i.e., 4 graph convolution layers as in~\cite{zhang2018end} with 32, 32, 32, and 1 units, 1 SortPooling layer with a threshold such that 60\% graphs have nodes less than the threshold, 2 1-D convolution layers (with 16 and 32 channels), and 1 dense layer (with 128 units). The learning rate is $10^{-4}$, the batch size is 50, the number of epochs is 50 with an early stopping round 20, and the hop number is 1. The dimensionality of the eigenspace is $d=128$ and we set $f(x)$ as the aforementioned normalization based on Frobenius norm.
\end{itemize}

\subsection{Hardware and Software Configurations}
All experiments are conducted on a server with the following configurations.
\begin{itemize}[leftmargin = 0.6cm]
\item Operating System: Ubuntu 18.04.1 LTS
\item CPU: Intel(R) Xeon(R) CPU E5-2699 v4 @ 2.20GHz
\item GPU: GeForce GTX TITAN X
\item Software: Python 3.6.9, TensorFlow 1.14.0, PyTorch 1.4.0, PyTorch Geometric 1.4.3, SciPy 1.4.1, NumPy 1.18.2, Cuda 10.1
\end{itemize}

\section{Proof of Theorem~\ref{thm1}}\label{sec:proofthm1}
\begin{proof}
Denote by $\mathbf{A}$ and $\mathbf{A}^\prime$, respectively, the adjacency matrices of $\pmb{G}$ and $\pmb{G}^\prime$. Since $\mathcal{E}(i,j) = \mathcal{E}^\prime(\mathcal{B}(i),\mathcal{B}(j))$, we have $\mathbf{A}_{i,j} = \mathbf{A}^\prime_{\mathcal{B}(i),\mathcal{B}(j)}$, $\mathcal{F}(\mathbf{A})_{i,j} = \mathcal{F}(\mathbf{A}^\prime)_{\mathcal{B}(i),\mathcal{B}(j)}$, and $\mathcal{G}(\mathbf{A})_{i,j} = \mathcal{G}(\mathbf{A}^\prime)_{\mathcal{B}(i),\mathcal{B}(j)}$, where $\mathcal{F}(\mathbf{A})$ and $\mathcal{G}(\mathbf{A})$ is the graph structure function in GNNs and the graph structure function in the eigenspace, defined in Eq.~\eqref{eq:GCN1} and Eq.~\eqref{eq:eigenGCN}, respectively.

Assume that the $l^{th}$ hidden layer in Eigen-GNN is permutation-equivariant, i.e., $\mathbf{H}^{(l)}_{i,:} = \mathbf{U}^{(l)}_{\mathcal{B}(i),:}, \forall 1\leq i \leq N$. Using Eq.~\eqref{eq:GCN1}, the $(l+1)^{th}$ layer is also permutation-equivariant:
\begin{equation}
\begin{aligned}
    \mathbf{H}^{(l+1)}_{i,:} & = \sigma(\sum\nolimits_j \mathcal{F}(\mathbf{A})_{i,j} \mathbf{H}^{(l)}_{j,:}\mathbf{W}^{(l)})
                             = \sigma(\sum\nolimits_j \mathcal{F}(\mathbf{A})_{i,j} \mathbf{U}^{(l)}_{\mathcal{B}(j),:} \mathbf{W}^{(l)}) \\
                             & = \sigma(\sum\nolimits_j \mathcal{F}(\mathbf{A}^\prime)_{\mathcal{B}(i),\mathcal{B}(j)} \mathbf{U}^{(l)}_{\mathcal{B}(j),:} \mathbf{W}^{(l)})
                               = \mathbf{U}^{(l+1)}_{\mathcal{B}(i),:}.
\end{aligned}
\end{equation}

Next, by induction, we only need to prove that the $0^{th}$ layer, i.e., $\mathbf{H}^{(0)}$ and $\mathbf{U}^{(0)}$, is also permutation-equivariant.
The node features are already permutation-equivariant by assumption. For the eigenbasis, since the top-$d$ eigenvalues of $\mathcal{G}(\mathbf{A})$ are unique, from linear algebra, we know that the eigenvectors are determined up to a sign for isomorphism graphs~\cite{babai1982isomorphism}.
Since we have adopted $f(x) = \left| x\right|$, we have $\mathbf{H}^{(0)}_{i,j} = \mathbf{U}^{(0)}_{\mathcal{B}(i),j}$. The theorem follows.
\end{proof}

\section{Proof of Theorem~\ref{thm2}}\label{sec:proofthm2}
\begin{proof}
Denote by $\mathbf{H}^{(l)}$ and $\mathbf{U}^{(l)}$, respectively, the representations of the nodes in the $l^{th}$ hidden layer in Eigen-GNN and SGC. Since the last layer in both models are task-specific, e.g., a softmax layer for node classification tasks or a readout function for graph classification tasks, we only need to prove that $\mathbf{U}^{\inf}$ converges to $\mathbf{H}^{(0)} = \mathbf{q}$, where $\mathbf{q} = \mathbf{Q}_{:,1}$ is a one-dimensional eigenbasis.

The hidden representations of SGC are calculated as follows~\cite{wu2019simplifying}.
\begin{equation}
    \mathbf{U}^{(l)} = \left( \tilde{\mathbf{D}}^{-\frac{1}{2}} \tilde{\mathbf{A}} \tilde{\mathbf{D}}^{-\frac{1}{2}} \right)^l \mathbf{X}.
\end{equation}

From linear algebra, the asymptotic behavior of $\mathbf{U}^{\inf}$ depends on the spectrum of $\tilde{\mathbf{D}}^{-\frac{1}{2}} \tilde{\mathbf{A}} \tilde{\mathbf{D}}^{-\frac{1}{2}}$. Denote by $\lambda_1 \leq \lambda_2 \leq ... \leq \lambda_N$ the eigenvalues of $\tilde{\mathbf{D}}^{-\frac{1}{2}} \tilde{\mathbf{A}} \tilde{\mathbf{D}}^{-\frac{1}{2}}$. From the spectral graph theory~\cite{ng2002spectral}, we have the following results.
\begin{itemize}[leftmargin = 0.6cm]
\item The spectrum is between $[-1,1]$, i.e., $-1 \leq \lambda_1 \leq \lambda_N \leq 1$.
\item If the graph is connected, $\lambda_N = 1$ and $\lambda_{N-1} < 1$.
\item $\lambda_1 = -1$ if and only if the graph is a bipartite.
\end{itemize}

Combining the above results, we have the following equation for a connected non-bipartite graph.
\begin{equation}
    -1 < \lambda_1 \leq ... \leq \lambda_{N-1} < \lambda_N = 1.
\end{equation}

Then, we know that $\mathbf{U}^{\inf}$ converges to the eigenvector corresponding to $\lambda_N$, which is also the eigenbasis corresponding to the largest absolute eigenvalue of $\tilde{\mathbf{D}}^{-\frac{1}{2}} \tilde{\mathbf{A}} \tilde{\mathbf{D}}^{-\frac{1}{2}}$, i.e.,
\begin{equation}
    \lim_{l\rightarrow \inf} \mathbf{U}^{(l)}_{:,i} = \mathbf{q}, \forall i.
\end{equation}
\end{proof}

\end{sloppy}
\end{document}